\documentclass[twoside]{article}

\usepackage[accepted]{aistats2020}

\usepackage[round]{natbib}

\usepackage[utf8]{inputenc} %
\usepackage[T1]{fontenc}    %
\usepackage[draft]{hyperref}
\usepackage{url}            %
\usepackage{booktabs}       %
\usepackage{amsfonts,amsmath}       %
\usepackage{nicefrac}       %
\usepackage{enumitem}       %
\usepackage{microtype,amsthm}      %
\usepackage{multirow}
\usepackage{xcolor}
\usepackage{algorithm,algorithmic}
\usepackage{graphicx,subcaption}

\newcommand{\mred}[1]{{\color{red} #1}}

\newcommand{\pmodel}[1]{q(#1;\theta)}
\newcommand{\pdata}[1]{p(#1)}
\newcommand{\norm}[1]{\lVert #1 \rVert}

\newcommand{\KL}[2]{\mathrm{KL}(#1\Vert #2)}
\newcommand{\mbf}[1]{\mathbf{#1}}
\newcommand{\mc}[1]{\mathcal{#1}}
\newcommand{\mrm}[1]{\mathrm{#1}}

\newcommand{\EE}{\mathbb{E}}

\newcommand{\bracket}[1]{\langle #1 \rangle}

\newcommand{\grad}{\mathrm{grad}}

\newcommand{\cH}{\mathcal{H}}
\newcommand{\cE}{\mathcal{E}}
\newcommand{\cF}{\mathcal{F}}

\newcommand{\cX}{\mathcal{X}}
\newcommand{\RR}{\mathbb{R}}

\newcommand{\KLp}{\mathrm{KL}_p}

\newcommand{\Wass}{\mc{P}}

\newcommand{\KLq}{\mrm{KL}_q}
\newcommand{\KLqh}{\mrm{KL}_{q^{1/2}}}

\newtheorem{remark}{Remark}[section]
\newtheorem{proposition}{Proposition}[section]
\newtheorem{lemma}{Lemma}[section]

\begin{document}

\runningtitle{A Wasserstein Minimum Velocity Approach to Learning Unnormalized Models}

\twocolumn[

\aistatstitle{A Wasserstein Minimum Velocity Approach to \\Learning Unnormalized Models}
\aistatsauthor{Ziyu Wang \And Shuyu Cheng \And Yueru Li \And Jun Zhu \And Bo Zhang}
\aistatsaddress{Dept. of Comp. Sci. \& Tech., BNRist Center, Institute for AI, THBI Lab, Tsinghua University}
]

\begin{abstract}
Score matching provides an effective approach to learning flexible unnormalized models, 
but its scalability is limited by the need to evaluate a second-order derivative. In this paper,
we present a scalable approximation to a general family of learning objectives including score matching, 
by observing a new connection between these objectives and Wasserstein gradient flows. 
We present applications with promise in learning neural density estimators on manifolds, and training implicit variational and Wasserstein auto-encoders with a manifold-valued prior. 
\end{abstract}

\section{INTRODUCTION}

A flexible approach to density estimation is to parameterize an unnormalized density function, or energy function. In particular, unnormalized models with energy parameterized by deep neural networks have been successfully applied to density estimation \citep{li2019learning,saremi2018deep} and  learning implicit auto-encoding models \citep{song2019sliced}.

Parameter estimation for such unnormalized models is highly non-trivial: the maximum likelihood objective is intractable, due to the presence of a normalization term. 
\emph{Score matching} \citep{hyvarinen2005estimation} is a popular alternative, %
yet applying score matching to complex unnormalized models can be difficult, as the objective involves the second-order derivative of the energy, rendering gradient-based optimization infeasible. 
In practice, people turn to scalable approximations of the score matching objective  
\citep{song2019sliced,hyvarinen2007connections,vincent2011connection,raphan2011least}, or other objectives such as the kernelized Stein discrepancy \citep[KSD; ][]{liu2016kernelized,liu2017learning}. %
So far, approximations to these objectives are developed on a case-by-case basis, leaving important applications unaddressed; for example, there is a lack of scalable learning methods for unnormalized models on manifolds~\citep{mardia2016score}. 

In this work, we present a unifying perspective to this problem, and derive scalable approximations for a variety of learning objectives including score matching. 
We start by interpreting these objectives as the initial \emph{velocity} of certain distribution-space gradient flows, which are simulated by common samplers. 
This novel interpretation leads to a scalable approximation algorithm for all such objectives, reminiscent to single-step contrastive divergence (CD-1). 

We refer to any objective with the above interpretation as above as a ``minimum velocity learning objective'', a term coined in the unpublished work \citep{movellan2007minimum}. \citet{movellan2007minimum} focused on the specific case of score matching; in contrast, 
our formulation generalizes theirs by lifting the concept of velocity from data space to distribution space, thus applies to different objectives as the choice of distribution space varies. 
For example, our method applies to score matching and Riemannian score matching when we choose the 2-Wasserstein space, and to KSD when we choose the $\cH$-Wasserstein space \citep{liu2017stein}; 
we can also derive instances of the minimum velocity learning objective when the distribution-space gradient flow corresponds to less well-studied samplers, such as \citep{zhang2018stochastic,lu2019accelerating}. %
Another gap we fill in is the development of a practically applicable algorithm, which we will discuss shortly.

Our algorithm is connected to previous work using CD-1 to estimate the \emph{gradient} of certain objectives \citep{hyvarinen2007connections,movellan2007minimum,liu2017learning}; %
however, there are important differences. 
From a theoretical perspective, we provide a unified derivation for all such objectives, including those not considered in previous work; our gradient-flow-based derivation is also simpler, and leads to an improved understanding of this approach. 
From an algorithmic perspective, we directly approximate the objective function instead of its gradient, enabling the use of regularization like early-stopping. 
More importantly, we identify an \emph{infinite-variance problem} in the approximate score matching objective, which has previously rendered the approximation impractical \citep{hyvarinen2007connections,saremi2018deep}; we further present a simple fix. 
As a side product of our work, our fix also applies to denoising score matching \citep{raphan2011least,vincent2011connection}, another score matching approximation that suffers from this problem. 

One important application of our method is in learning unnormalized models on manifolds, as our method leads to a scalable approximation for the Riemannian score matching objective. Density estimation on manifolds 
is needed in areas such as image analysis \citep{srivastava2007riemannian}, geology \citep{davis1986statistics} and bioinformatics \citep{boomsma2008generative}.
Moreover, our approximation leads to flexible inference schemes for variational and Wasserstein auto-encoders with manifold-valued latent variables, as it enables gradient estimation for implicit variational distributions on manifolds. 
Auto-encoders with a manifold-valued latent space %
can capture the distribution of certain types of data better. For example, a hyperbolic latent space could be more suitable when the data has a hierarchical structure \citep{mathieu2019hierarchical,anonymous2020poincar}, and a hyper-spherical prior could be more suitable for directional data \citep{davidson2018hyperspherical}. As we shall see in experiments, our method improves the performance of manifold-latent VAEs and WAEs.

The rest of this paper is organized as follows: Section~2  reviews the preliminary knowledge: manifolds%
, gradient flows and their connection to common sampling algorithms. We present our method in Section~3 and its applications in Section~4. Section~5 contains a review of the related work, and Section~6 contains experiments. We provide our conclusions in Section~7.

\section{PRELIMINARIES}

\subsection{Manifolds, Flows and the 2-Wasserstein Space}

We recall concepts from differential manifolds that will be needed below.

A (differential) \emph{manifold} $\mc{M}$ is a topological space locally diffeomorphic to an Euclidean or Hilbert space. A manifold is covered by a set of \emph{charts}, which enables the use of coordinates locally, and specifies a set of basis $\{\partial_i\}$ in the local tangent space. A \emph{Riemannian manifold} further possesses a \emph{Riemannian structure}, which assigns to each tangent space $\mc{T}_p \mc{M}$ an inner product structure. The Riemannian structure can be described using coordinates w.r.t. local charts.

The manifold structure enables us to differentiate a function along curves. Specifically, consider a curve $c:[0,T]\mapsto \mc{M}$, and a smooth function $f:\mc{M}\mapsto \mathbb{R}$. At $c(t)\in \mc{M}$, a \emph{tangent vector} $\left.\frac{dc}{dt}\right|_t\in \mc{T}_{c(t)}\mc{M}$ describes the velocity of $c$ passing $c(t)$; the \emph{differential} of the function $f$ at $c(t)$, denoted as $(df)_{c(t)}$, is a \emph{linear} map from $\mc{T}_{c(t)}\mc{M}$ to $\mathbb{R}$, such that for all $c$
$$
(df)_{c(t_0)}\left(\left.\frac{dc}{dt}\right|_{t_0}\right) = 
\left.\frac{d}{dt} f(c(t))\right|_{t_0}.
$$
A \emph{tangent vector field} assigns to each $p\in \mc{M}$ a tangent vector $V_p\in \mc{T}_p\mc{M}$. 
It determines a \emph{flow}, a set of curves $\{\phi_p(t):p\in \mc{M}\}$ which all have $V_{\phi_p(t)}$ as their velocity. 
On Riemannian manifolds, the \emph{gradient} of a smooth function $f$ is a tangent vector field $p\mapsto \grad_p f$ such that $\bracket{\grad_p f, v}=(df)_p(v)$ for all $v\in T_p\mc{M}$. 
It determines the \emph{gradient flow}, which generalizes the Euclidean-space notion $dx=\nabla_x f(x) dt$. 

We will work with two types of manifolds: the data space $\cX$ when we apply our method to manifold-valued data, and the space of probability distributions over $\cX$. 
On the space of distributions, we are mostly interested in the \emph{2-Wasserstein space} $\Wass(\cX)$, a Riemannian manifold. 
The following properties of $\Wass(\cX)$ will be useful for our purposes \citep{villani2008optimal}:
\begin{enumerate}[leftmargin=*]
\item Its tangent space $\mc{T}_p\Wass(\cX)$ can be identified as a subspace of the space of vector fields on $\cX$; the Riemannian metric of $\Wass(\cX)$ is defined as
\begin{equation}\label{eq:riem-metric-w2}
    \bracket{X,Y}_p:=\EE_{p(u)}\bracket{X(u),Y(u)}_{u},
\end{equation}
for all $p\in\Wass(\cX),X,Y\in\mc{T}_p\Wass(\cX)$; 
the inner product on the right hand side above is determined by the Riemannian structure of $\cX$.
\item The gradient of the KL divergence functional $\KLp(q):=\KL{q}{p}$ in $\Wass(\cX)$ is
\begin{equation}\label{eq:w2-kld-grad}
    (\grad_q \KLp)(u)=\grad_u \log \frac{q(u)}{p(u)}.
\end{equation}
\end{enumerate}
We will also consider a few other spaces of distributions, including the Wasserstein-Fisher-Rao space \citep{lu2019accelerating}, and the $\cH$-Wasserstein space introduced in \citep{liu2017stein}.

On the data space, %
we need to introduce the notion of density, i.e. the Radon–Nikodym derivative w.r.t. a suitable base measure. The Hausdorff measure is one such choice; it reduces to the Lebesgue measure when $\cX=\RR^n$. In most cases, distributions on manifolds are specified using their density w.r.t. the Hausdorff measure; e.g. ``uniform'' distributions has constant densities in this sense. 

Finally, the data space $\cX$ will be embedded in $\mathbb{R}^n$; we refer to real-valued functions on the space of distributions as functionals; we denote the functional $q\mapsto \KL{q}{p}$ as $\KLp$; we adopt the Einstein summation convention, and omit the summation symbol when an index appears both as subscript and superscript on one side of an equation, e.g. $v^i\partial_i:=\sum_i v^i\partial_i$.

\subsection{Posterior Sampling by Simulation of Gradient Flows}\label{sec:background-samplers}

Now we review the sampling algorithms considered in this work. They include diffusion-based MCMC, particle-based variational inference, and other stochastic interacting particle systems.

\paragraph{Riemannian Langevin Dynamics} Suppose our target distribution has density $p(x)$ w.r.t. the Hausdorff measure of $\cX$. 
In a local chart $U\subset \cX$, let $G: U\rightarrow \RR^{m\times m}$ be the coordinate matrix of its Riemannian metric. 
Then the \emph{Riemannian Langevin dynamics} corresponds to the following stochastic differential equation in the chart\footnote{
    \eqref{eq:riem-ld} differs from definitions in some works \citep[e.g.][]{ma2015complete}. This is because we define $p$ as the density w.r.t. the Hausdorff measure of $\cX$, while they use the Lebesgue measure. See also \citep{xifara2014langevin,hsu2008brief}.
}:
\begin{equation}\label{eq:riem-ld}
dx = V(x)dt + \sqrt{2G^{-1}(x)}dB_t
\end{equation}
 where 
\begin{align}
    V^i(x) =& g^{ij}\partial_j\left(\log p(x)-\frac{\log |G(x)|}{2}\right) + \partial_j g^{ij},
\end{align}
and $(g^{ij})$ is the coordinate of the matrix $G^{-1}$.
It is known \citep{villani2008optimal} that the Riemannian Langevin dynamics is the gradient flow of the KL functional $\KLp(q):=\KL{q}{p}$ in the 2-Wasserstein space $\Wass(\cX)$. %

\paragraph{Particle-based Samplers} A range of samplers approximate the gradient flow of $\KLp$ in various spaces, using deterministic or stochastic interacting particle systems.\footnote{
    There are other particle-based samplers \citep{liu2019understanding,liu2019understandingA,taghvaei19accelerated}
    corresponding to accelerated gradient flows. However, as we will be interested in the initial velocity of the flow, they do not lead to new MVL objectives. 
} 
For instance, Stein variational gradient descent \citep[SVGD;][]{liu2016stein} simulates the gradient flow in the so-called $\cH$-Wasserstein space \citep{liu2017stein}, which replaces the Riemannian structure in $\Wass(\cX)$ with the RKHS inner product. Birth-death accelerated Langevin dynamics \citep{lu2019accelerating} is a stochastic interacting particle system that simulates to the gradient flow of $\KLp$ in the Wasserstein-Fisher-Rao space. Finally, the stochastic particle-optimization sampler  \citep[SPOS;][]{zhang2018stochastic,chen2018unified} combines the dynamics of SVGD and Langevin dynamics; as we will show in Appendix~\ref{suppl:spos-gf}, SPOS also has a gradient flow structure. 

\section{WASSERSTEIN MINIMUM VELOCITY LEARNING}

In this section, we present our framework, which concerns all learning objectives of the following form:
\begin{equation}\label{eq:mvl-defn-again}
L_{\mrm{mvl}}(\theta) := -\left.\frac{d}{dt} \KL{p_t}{q_\theta}\right|_{t=0},
\end{equation}
where $p_t$ is defined as the gradient flow of $\KLq$ in a suitable space of probability measures (e.g. the 2-Wasserstein space). 
We refer to any such objective as a ``minimum velocity learning (MVL) objective''; as we shall see below, $L_{\mrm{mvl}}$ equals the initial \emph{velocity} of the gradient flow $\norm{\grad_p\KLq}^2$, in the corresponding distribution space. 

In the following subsections, we will first set up the problem, and motivate the use of \eqref{eq:mvl-defn-again} by connecting it to score matching; 
then we present our approximation to \eqref{eq:mvl-defn-again}, and its variance-reduced version; 
we also address the infinite-variance issue in two previous approximators for the score matching objective. 
Finally, we briefly discuss other instances of the MVL objective that our method can be applied to.

\subsection{Score Matching and a Wasserstein Space View}\label{sec:score-matching-and-mvl}

Consider parameter estimation in the unnormalized model $q(x;\theta) := \frac{1}{Z(\theta)}\exp(-\cE(x;\theta))$. 
Maximum likelihood estimation is intractable, due to the presence of the normalizing constant $Z(\theta)=\int\exp(-\cE(x;\theta))dx$. 
Score matching %
circumvents this issue by minimizing the Fisher divergence %
\begin{align}
D_F(p|q):=\:&
\frac{1}{2}\EE_{\pdata{x}}\left[\norm{\nabla_x \log\pdata{x}-\nabla_x \log\pmodel{x}}^2\right],\label{eq:fisher-div}
\end{align}
which does not depend on the normalization constant. 
While \eqref{eq:fisher-div} involves the unknown $\nabla_x \log \pdata{x}$ term, \citet{hyvarinen2005estimation} shows that it equals
\begin{align}
\EE_{\pdata{x}}\left[\Delta\log\pmodel{x}+\frac{1}{2}\norm{\nabla\log\pmodel{x}}^2\right],\label{eq:fisher-div-transformed}
\end{align}
plus a constant independent of $\theta$. 
Thus we can estimate the Fisher divergence at the cost of introducing a second-order derivative.

Unfortunately, optimization w.r.t. second-order derivatives is prohibitively expensive when the energy is parameterized by deep neural networks, and scalable approximation to the score matching objective must be developed. Our work starts by observing
$$
D_{F}(p|q) = \frac{1}{2}\norm{\grad_p\KLq}^2,
$$
where the gradient and norm are defined in $\Wass(\cX)$, and the manifold $\cX$ inherits the Riemannian metric from $\RR^n$. This follows directly from \eqref{eq:riem-metric-w2}-\eqref{eq:w2-kld-grad}.

Now let $p_t$ be the gradient flow of $\KLq$, i.e. $\frac{d}{dt}p_t=-\grad_{p_t}\KLq$. Then %
\begin{align}
\norm{\grad_p\KLq}^2 &= d(\KLq)_p(\grad_p\KLq) \nonumber \\
&= -\left.\frac{d}{dt} \KL{p_t}{q_\theta}\right|_{t=0} = L_\mrm{mvl}(\theta).\label{eq:rearrange-term-0}
\end{align}
Therefore, score matching is a special case of the MVL objective \eqref{eq:mvl-defn-again}, when the space of distributions is chosen as $\Wass(\cX)$.

\subsection{Approximating the MVL Objective}\label{sec:the-mvl-approx}

While the MVL objective has a closed-form expression, it usually involves second-order derivatives. %
In this subsection, we will derive an efficient approximation scheme for the MVL objective. Our approximation will only involve first-order derivatives, thus it can be easily implemented using automatic differentiation softwares (e.g. TensorFlow).

First, observe that \eqref{eq:rearrange-term-0}
holds regardless of the chosen space of distributions. 
Denote $\cH[p] := \EE_p \log p$, $\cF[p] := \EE_p \log q = -\EE_p \cE$, so $\KLq = \cH - \cF$, then we can transform the above into
\begin{align}
     &\bracket{\grad_p\cH-\grad_p\cF,\grad_p\cH-\grad_p\cF} \nonumber\\
    =&\norm{\grad_p\cH}^2 - 
        2\left\langle \grad_p \cF, \grad_p\cH-\frac{1}{2}\grad_p\cF \right\rangle \nonumber\\
    =& \norm{\grad_p\cH}^2 - 2\bracket{\grad_p \cF, \grad_p\KLqh} \label{eq:tractable-obj}.
\end{align}
As the first term in \eqref{eq:tractable-obj} is independent of $\theta$, the MVL objective is always equivalent to the second term. We will approximate the second term by simulating a modified gradient flow: let $\tilde{p}_t$ be the distribution obtained by running the sampler targeting $q^\mred{1/2}$. Then
\begin{align}
    \bracket{\grad_p \cF, -\grad_p\KLqh} = 
        (d\cF)_p(-\grad_p\KLqh)\nonumber \\
        =
    \lim_{\epsilon\rightarrow 0} \frac{\EE_{\tilde{p}_t}\log q_\theta - \EE_{p}\log q_\theta}{\epsilon}.\label{eq:tractable-obj-transformed}
\end{align}
\eqref{eq:tractable-obj-transformed} can be approximated by replacing the limit with a fixed $\epsilon$, and running the corresponding sampler starting from a mini-batch of training data. 
The approximation becomes unbiased when $\epsilon\rightarrow 0$.

\subsubsection{A Control Variate}\label{sec:cv}

We have derived an estimator of \eqref{eq:tractable-obj-transformed} with vanishing bias. 
However, the estimator will suffer from high variance when the sampler used in the MVL objective consists of It\^o diffusion. Fortunately, we can solve this problem with a control variate.

To illustrate the problem as well as our solution, suppose $\{p_t\}$ corresponds to Langevin dynamics, and (without loss of generality) we use a batch size of $1$ in estimation. Our estimator is then %
\begin{equation*}%
\hat{L}_\epsilon = \frac{2}{\epsilon}\left[\mc{E}(x^+) - \mc{E}\left(x^+ - \epsilon\nabla_x \frac{\cE(x^+)}{2} + \sqrt{2\epsilon} Z\right)\right],
\end{equation*}
where $x^+$ is sampled from the training data, and $Z\sim \mc{N}(0,I)$. 
By Taylor expansion\footnote{We need to expand to the second order when the increment is a discretization of some It\^o diffusion.}, $\hat{L}_\epsilon/2$ equals%
\begin{align}
 &\frac{1}{2}\norm{\nabla_x \cE(x^+)}^2 - Z^\top \nabla^2_x \cE(x^+) Z
 - \sqrt{\frac{2}{\epsilon}}Z^\top \nabla_x \cE(x^+) \nonumber\\
 &+ o(1),\label{eq:taylor-exp}
\end{align}%
and as $\epsilon\rightarrow 0$,
$
\mathrm{Var}\:\hat{L}_\epsilon = \Theta\left({\epsilon}^{-1}\right)\rightarrow \infty.
$

Now we can see the need for a control variate. In this LD example, the control variate $\sqrt{2/\epsilon}Z^\top \nabla_x \cE(x^+)$ will remove the infinite-variance term; 
More generally, our control variate is always the inner product of $\nabla_x\cE(x^+)$ and the diffusion term in the sampler.

Wrapping up, our approximate MVL objective is calculated as follows:%
\begin{enumerate}[leftmargin=*]
    \item Sample a mini-batch of input $\{x_i\}_{i=1}^B$.%
    \item Run a single step of the sampling algorithm on $\{x_i\}$ targeting $q_{1/2}\propto \exp(-\cE(x;\theta)/2)$, with a step-size of $\epsilon$. Denote the resulted state as $\{x^-_i\}$.
    \item Return $\frac{1}{B}\sum_{j=1}^B(\cE(x_i)-\cE(x^-_i))$ plus the control variate.
\end{enumerate}
The approximation becomes unbiased as $\epsilon\rightarrow 0$, and has $O(1)$ variance\footnote{
    under mild assumptions controlling the growth of $\nabla^3\cE$ (e.g. bounded by a polynomial), so that the residual term in \eqref{eq:taylor-exp} will have bounded variance when averaged over $Z$.%
} regardless of $\epsilon$.

\subsection{On CD-1 and Denoising Score Matching: Pitfalls and Fixes}\label{sec:longest-name}

As a side product, we show that our variance analysis explains the pitfall of two well-known approximations to the score matching objective: 
CD-1 \citep{hyvarinen2007connections} and denoising score matching \citep[DSM]{vincent2011connection}. 
Both approximations become unbiased as a step-size hyper-parameter $\epsilon\rightarrow 0$, but did not match the performance of exact score matching in practice, as witnessed in \citet{hyvarinen2007connections,saremi2018deep,song2019sliced}. 
We propose novel control variates for these approximators. As we will show in Section~\ref{sec:synth}, the variance-reduced versions of the approximations have comparable performance to the exact score matching objective. 

\paragraph{DSM} DSM considers the objective
\begin{equation}\label{eq:dsm}
L_\mrm{dsm}(\theta) = \EE_{p(x)\mc{N}(z|0,I)}\norm{x+\sigma z-(x+\psi_\theta(x+\sigma z))}^2.
\end{equation}
The first two terms inside the norm represent a noise corrupted sample, and $\psi_\theta$ represents a ``single-step denoising direction'' \citep{raphan2011least}. It is proved that the optimal $\psi$ satisfies $\psi=\sigma^2 \nabla \log \tilde{p}$, where $\tilde{p}$ is the density of the corrupted distribution \citep{raphan2011least,vincent2011connection}. 

Consider the stochastic estimator of \eqref{eq:dsm}. We assume a batch size of $1$, and denote the data sample as $x$. To keep notations consistent, denote $\epsilon=\sigma^2$, $\psi_\theta(x)=\epsilon \nabla_x \cE(x;\theta)$. Then the estimator is
$$
\hat{L}_\mrm{dsm} 
= \norm{x+\sqrt{\epsilon} z-\epsilon\nabla_x\cE(x+\sqrt{\epsilon} z;\theta)-x}^2. 
$$
As is similar to Section~\ref{sec:cv}, we can show by Taylor expansion (see Appendix~\ref{suppl:deriv-dsm-variance}) that
\begin{align}
\lim_{\epsilon\rightarrow 0}\epsilon^{-2}\EE\hat{L}_\mrm{dsm} &= 2D_\mrm{Fisher}(p|q)+\mrm{(const)}, \label{eq:dsm-stoc-final-s}\\
\lim_{\epsilon\rightarrow 0}\epsilon^{-2}\mrm{Var}\:\hat{L}_\mrm{dsm} &= \infty;
\end{align}
furthermore, the variance reduced objective
\begin{equation} \label{eq:dsm-stoc-final-e}
    \frac{\hat{L}_\mrm{dsm}-(\epsilon\norm{z}^2-2\epsilon^{3/2}z^\top\nabla\cE(x))}{\epsilon^2}
\end{equation}
is unbiased with finite variance.

\paragraph{CD-1 with Langevin Dynamics}
Proposed as an approximation to the maximum likelihood estimate, the $K$-step contrastive divergence (CD-$K$) learning rule updates the model parameter with
\begin{equation}\label{eq:cd-learning-rule}
    \theta_{\ell+1} \gets \theta_\ell + \nu \left[
    \EE_p \partial_\theta \cE - 
    \EE_{p_K} \partial_\theta \cE
    \right],
\end{equation}
where $\nu$ is the learning rate, and $p_K$ is obtained from $p$ by running $K$ steps of MCMC. 
\eqref{eq:cd-learning-rule} does not define a valid objective, since $p_K$ also depends on $\theta$; however, \citet{hyvarinen2007connections} proved that when $K=1$ and the sampler is the Langevin dynamics, \eqref{eq:cd-learning-rule} recovers the \emph{gradient} of the score matching objective. 

Using the same derivation as in Section~\ref{sec:cv}, we can see that as the step-size of the sampler approaches 0 (and $\nu$ is re-scaled appropriately), the gradient produced by CD-$1$ also suffers from infinite variance, and this can be fixed using the same control variate. %

However, practical utility of CD-1 is still hindered by the fact that it does not correspond to a valid learning \emph{objective}; consequently, it is impossible to monitor the training process for CD-1, or introduce regularizations such as early stopping\footnote{
    In practice, the term $\EE_p\cE-\EE_{p_K}\cE$ is often used to tract the training process of CD-$K$. 
    It is not a proper loss; we can see from \eqref{eq:tractable-obj} that when $K=1$ and $\epsilon\rightarrow 0$, $\EE_p\cE-\EE_{p_K}\cE$ is significantly different from the proper score matching (MVL) loss, by a term of $\frac{1}{2}\norm{\grad_p\cF}^2$.
}.

\subsection{Instances of MVL Objectives}\label{sec:other-objs}

As the previous derivation is independent of the distribution space of choice, we can derive approximations to other learning objectives using samplers other than LD. %
An important example is the Riemannian score matching objective, which corresponds to Riemannian LD; we will discuss it in detail in Section~\ref{sec:mvl-riem-manif}. 
Another example is when we choose the sampler as SVGD. In this case, we will obtain an approximation to the kernelized Stein discrepancy, generalizing the derivation in \citep{liu2017learning}. 
When the sampling algorithm is chosen as SPOS, the corresponding MVL objective will be an interpolation between KSD and the Fisher divergence. 
See Appendix~\ref{suppl:mvl-spos-ksd} for derivations. 
Finally, the use of birth-death accelerated Langevin dynamics leads to a novel learning objective. 

In terms of applications, our work focuses on learning neural energy-based models, and these objectives do not improve over score matching in this aspect. 
However, these derivations are useful since they generalize previous discussions, and establish new connections between sampling algorithms and learning objectives. 
It is also possible that these approximate objectives could be useful in other scenarios, such as learning kernel exponential family models \citep{sriperumbudur2017density}, %
improving the training of GANs \citep{liu2017learning} or amortized variational inference methods \citep{ruiz2019contrastive}.

\section{APPLICATIONS}\label{sec:appl-autoencoders}

We now present applications of our work, including a scalable learning algorithm for unnormalized models on manifolds, as well as its application on learning implicit auto-encoders with manifold-valued priors. 

\subsection{MVL on Riemannian Manifolds}\label{sec:mvl-riem-manif}

Density estimation on manifolds is needed in many application areas. %
While it is natural to consider unnoramlized models on manifolds, there has been a lack of scalable learning methods. Here we address this issue, by applying our method to obtain a scalable approximation to the Riemannian score matching objective \citep{mardia2016score}.%

Given the data manifold $\cX$, we define an unnormalized model on it by parameterizing the log density w.r.t. the Hausdorff measure, and define the density as $q(x;\theta)\propto\exp(-\cE(x;\theta))$. The Riemannian score matching objective will have the same form as \eqref{eq:fisher-div}; although the norm in \eqref{eq:fisher-div} is now determined by the metric on $\cX$, and the base measure of the densities has changed.

It is easy to verify that the derivation in Section~\ref{sec:score-matching-and-mvl} still applies in the manifold case. Thus, the Riemannian score matching objective is a special case of the MVL objective, in which the distribution space is still chosen as $\Wass(\cX)$. 
The difference is that $\Wass(\cX)$ is now defined with the non-trivial data-space metric, and 
the gradient flow of $\KLq$ becomes the Riemannian Langevin dynamics \eqref{eq:riem-ld}. We can approximate the objective by doing a single step of Riemannian LD for small $\epsilon$:
\begin{align}\label{eq:loss-mvl-rld}
\vspace{-0.5em}
L_\text{mvl-rld} =  \frac{2}{\epsilon}\big(
    {\cE}(y^-;\theta)-{\cE}(y;\theta)-\underbrace{\sqrt{2\epsilon}\partial_i {\cE}(y)z^i}_{\text{control variate}}
\big).
\end{align}
In \eqref{eq:loss-mvl-rld}, $y$ is the local coordinates of a sampled data point, $G(y)$ is the Riemannian metric, and $y^-$ is obtained by running Riemannian Langevin dynamics\footnote{
While readers familiar with Riemannian Brownian motion may notice that \eqref{eq:riem-ld-again} is only defined before the particle escapes the local chart, 
this is good enough for our purpose: we are only concerned with infinitesimal time, and escape probability approaches $0$ as $\epsilon\rightarrow 0$. 
See Appendix~\ref{suppl:local-coord-validity}.%
} targeting $q_{1/2}$:
\begin{align}
(y^-)^i =\; & y^i + \epsilon \left(
    -g^{ij}\partial_j\frac{\cE(y;\theta)+\log |G(y)|}{2} + \partial_k g^{ik}
\right) \nonumber \\& +\sqrt{2\epsilon}z^i,\label{eq:riem-ld-again}\\
z \sim\; & \mc{N}(0,G^{-1}(y)).\nonumber
\end{align}

\subsection{Learning Implicit AEs with Manifold Prior}\label{sec:appl-ae-manif}

Recently, there is a surge of interest in auto-encoding models with manifold-valued priors. 
In this section, we present a new training method for implicit auto-encoders with manifold priors, based on the above Riemannian score matching algorithm. %

Formally, auto-encoders model the observed data by marginalizing out a latent code variable, $p(x;\theta):=\int p(x|z;\theta)p(z)dz$. To enable tractable learning, they define an additional ``encoder'' distribution $q(z|x;\phi)$. We will consider two types of auto-encoders:
\begin{enumerate}[leftmargin=*]
    \item VAEs with implicit encoder, which maximizes %
    $
    \EE_{p(x)}\EE_{q(z|x;\phi)}\log\frac{p(z)p(x|z;\theta)}{q(z|x;\phi)},
    $
    the evidence lower bound.
    $q$ is a reparameterized implicit distribution, i.e. for fixed $x$, $q(z|x;\phi)$ is defined as the pushforward measure of a simple distribution $q_0(\epsilon)$, by a DNN that takes $x$ and $\epsilon$ as input.
    \item Wasserstein auto-encoders (WAEs), which minimizes the 1-Wasserstein distance between the model and data distributions by minimizing
    $
        \EE_{p(x)}\left[
            \EE_{q(z|x;\phi)}c(G(z;\theta),x)
        \right] + \lambda D(\tilde{q}(z), p(z)),
    $
    where $G(z;\theta)$ is the deterministic decoder, i.e. $p(x|z;\theta)=\delta(x-G(z;\theta))$; $c$ is a user-specified reconstruction error, $\tilde{q}(z) := \int p(x)q(z|x)dx$ is the aggregated prior, $\lambda$ is a hyperparamter, and $D$ is an arbitrary divergence. We use the exclusive KL divergence as $D$.
\end{enumerate}

Both objectives are intractable, as they include the entropy of a latent-space distribution with intractable density: $H[q(z|x;\phi)]$ for VAE, and $H[\tilde{q}(z)]$ for WAE. 
However, %
it is known that to obtain $\nabla_\phi H(q)$, it suffices to estimate the score function $\nabla_z \log q(z)$. Specifically, let $q(z;\phi)$ be the pushforward of $p(\epsilon)$ by $f_q(\epsilon;\phi)$. Then we have%
\begin{equation}\label{eq:ent-est-to-score-est}
\nabla_\phi H[q(z)] = -\EE_{\epsilon}\left[\nabla_z\log q(z) \nabla_\phi f_q(\epsilon;\phi) \right].
\end{equation}
Score estimation can be done by fitting an unnormalized model $\cE(z)$ on the distribution $q(z)$, and approximating $\nabla_z \log q(z)$ above with $-\nabla_z \cE(z)$. (For VAE, we will fit a conditional unnormalized model to approximate the conditional entropy.)

A variant of this idea is explored in \citet{song2019sliced}, and outperforms existing learning algorithms for implicit AEs. As argued by \citep{shi2018spectral,li2018gradient}, this method is advantageous as it directly estimates the \emph{score function} of the latent-space distribution, instead of obtaining gradient from density (ratio) estimations; the latter could lead to arbitrary variations in the gradient estimate. 

When the latent variables are defined on an embedded manifold (e.g. hyper-spheres), we can no longer use the Euclidean score estimators to approximate the learning objective, as the entropy of the latent-space distribution w.r.t. the Lebesgue measure is usually undefined.  
However, we can still approximate the objective by doing score estimation inside the manifold: let $q(z)$ be the density w.r.t. the Hausdorff measure, and $H$ be the corresponding relative entropy functional. Then \eqref{eq:ent-est-to-score-est} will still hold; see Appendix~\ref{suppl:deriv-ent-score-est}. 
We can estimate the score function in \eqref{eq:ent-est-to-score-est} by with an unnormalized model on manifold, learned with the objective \eqref{eq:loss-mvl-rld}. 

Wrapping up, we obtain an efficient algorithm to train auto-encoders with a manifold-valued prior.

\section{RELATED WORK}\label{sec:related-work}

Our work concerns scalable learning algorithms for unnormalized models. This is a longstanding problem in literature, and some of the previous work is discussed in Section~1. Other notable work includes noise contrastive estimation \citep{gutmann2010noise} and Parzen score matching \citep{raphan2011least}. However, to our knowledge, they have not been applied to complex unnormalized models parameterized by DNNs.

Apart from the MVL formulation used in this work, there exists other work on the connection between learning objectives of unnormalized model and infinitesimal actions of sampling dynamics (or other processes): 
\begin{itemize}[leftmargin=*]\vspace{-0.3em}
    \item The \emph{minimum probability flow framework} \citep{sohl2011minimum} studies the slightly different objective $\lim_{\epsilon\rightarrow 0}\frac{1}{\epsilon}\KL{p_0}{p_\epsilon}$, where $\{p_t\}$ is the trajectory of the sampler. It recovers score matching as a special instance, and leads to a tractable learning objective for discrete models.
    \item Many of the objective functions we have considered are also instances of the \emph{Stein discrepancy}. This interpretation is helpful in establishing theoretical properties \citep{gorham19measuring} and deriving new objectives \citep{aless2019minimum}. %
    \item \citet{lyu2009interpretation} observes a different connection between score matching and (derivative of) KL divergence; specifically they showed $D_F(p|q) = \left.\frac{d}{dt}\mrm{KL}(p_t|q_t)\right|_{t=0}$, where $\{p_t\},\{q_t\}$ are obtained by doing \emph{Brownian motion} starting from $p$ or $q$. %
\end{itemize}
\vspace{-0.3em}
As those formulations have different motivations compared with ours, they 
do not lead to scalable learning objectives for continuous models. 

\section{EVALUATION}

\subsection{Synthetic Experiments}\label{sec:synth}

To demonstrate the proposed estimators have small bias and variance, we first evaluate them on low-dimensional synthetic data. We will also verify %
that our control variate in Section~\ref{sec:longest-name} improves the performance of CD-1 and DSM.

\subsubsection{Approximations to Score Matching}\label{sec:exp-score-matching-obj}

In this section, we evaluate our MVL approximation to the Euclidean score matching objective \eqref{eq:fisher-div-transformed}, %
as well as the variance-reduced DSM objective. An experiment evaluating the variance-reduced CD-1 objective is presented in Appendix~\ref{suppl:cd-cv}.

We evaluate the bias and variance of our estimators by comparing them to sliced score matching (SSM), an unbiased estimator for \eqref{eq:fisher-div-transformed}. 
We choose the data distribution $p$ as the 2-D banana dataset from \citet{wenliang2018learning}, and the model distribution $q_\theta$ as an EBM trained on that dataset. 
We estimate the squared bias with a stochastic upper bound using $5\times 10^6$ samples; see Appendix~\ref{suppl:exp-details-smobj} for details.

\begin{figure}[htbp]
    \centering
    \includegraphics[width=\linewidth]{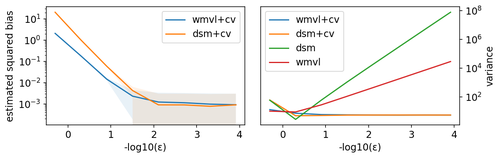}
    \caption{Estimated squared bias (left) and variance (right) of the approximate score matching objectives with varying step-size.}
    \vspace{-0.6em}
    \label{fig:bias-and-var}
\end{figure}

The results are shown in Figure~\ref{fig:bias-and-var}. %
We can see that for both estimators, the bias is negligible at $\epsilon\le 10^{-2}$. We further use a z-test to compare the mean of the two estimators (for $\epsilon=6\times 10^{-5}$) with the mean of SSM. The p value is $0.48$ for our estimator and $0.19$ for DSM, indicating there is no significant difference in either case. 
The variance of the estimators, with and without our control variate, are shown in Fig.\ref{fig:bias-and-var} right. As expected, the variance grows unbounded in absence of the control variate, and is approximately constant when it is added. From the scale of the variance, we can see that it is exactly this variance problem that causes the failure of the original DSM estimator.%

\subsubsection{Density Estimation on Manifolds}\label{sec:exp-riem-sm}

\begin{figure}[htbp]
    \vspace{-0.3em}
    \centering
    \includegraphics[width=0.85\linewidth]{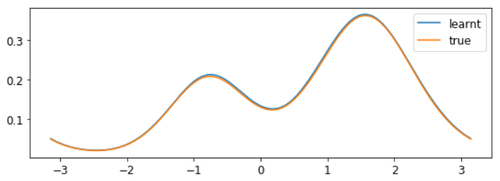}
   \vspace{-0.5em}
    \caption{Density estimation on $\mc{S}^1$: learned energy vs ground truth in polar coordinates.}\label{fig:synth-S1}
   \vspace{-0.7em}
\end{figure}

We now evaluate our approximation to the Riemannian score matching objective, by learning neural energy-based models on $\mc{S}^1$ and $\mc{S}^2$. The target distributions are mixtures of von-Mises-Fisher distributions. 
In Figure~\ref{fig:synth-S1}, we plot the log densities of the ground truth distribution as well as the learned model on $\mc{S}^1$. We can see the two functions matches closely, suggesting our method is suitable for density estimation on manifolds. 
Results on $\mc{S}^2$ are similar and will be presented in \ref{suppl:s2}; detailed setups are deferred to Appendix~\ref{suppl:exp-details-smobj}.

\subsection{Implicit AEs with Manifold Prior}\label{sec:exp-ae}
\vspace{-0.2em}

We now apply our method to train implicit auto-encoding models with manifold-valued prior. %
Experiment setups mainly follow \citet{song2019sliced}; see  Appendix~\ref{suppl:exp-ae}.

Note that there is an important difference from \citet{song2019sliced} in our implementation: for (conditional) score estimation, we parameterize an scalar energy function $\mc{E}_\theta(z)$ and use $\nabla_z\mc{E}_\theta(z)$ as the score estimate, while 
\citet{song2019sliced} directly parameterize a vector-value network $\psi_\theta(z)$. 
Since directly using a feed-forward network (FFN) for $\cE$ does not work well in practice, 
we parameterize the energy function as $\cE_\theta(z):=z^\top\psi_\theta(z)$, %
where $\psi$ is parameterized in the same way as \citet{song2019sliced}.
This can be seen as correcting an initial score approximation $\psi$ to make it conservative.
In addition to being conceptually desirable (as score functions are conservative fields), 
this approach leads to significant improvements in the WAE experiments.

\vspace{-0.3em}
\subsubsection{Implicit VAEs}
\vspace{-0.2em}

We apply our method to train hyperspherical VAEs \citep{davidson2018hyperspherical} with implicit encoders on the MNIST dataset. %
Our encoder and decoder architecture follows \citet{song2019sliced}, with the exception that we normalize $z$ so it lies on $\mc{S}^n$. 

We consider $n_z\in\{8,32\}$. Baseline methods include hyperspherical VAE with explicit encoders and Euclidean VAEs. We report the test log likelihood estimated with annealed importance sampling \citep{wu2016quantitative,neal2001annealed}, as well as its standard deviation across 10 runs. 

\newcommand{\wsd}[1]{{\tiny$\pm #1$}}
\begin{table}[htb]\centering
\setlength{\tabcolsep}{3pt}
    \vspace{-0.3em}
\begin{tabular}{ccccc}
\toprule
\multirow{2}{*}{} & \multicolumn{2}{c}{$n_z=8$} & \multicolumn{2}{c}{$n_z=32$} \\
                        & Euc.  & Sph.      & Euc.   & Sph.       \\ \midrule
Exp.                & 96.45\wsd{0.10}      & 95.47\wsd{0.08}          & 90.28\wsd{0.37}       & 91.32\wsd{0.07}           \\
Imp.                & 95.84\wsd{0.19}      & \textbf{94.72}\wsd{0.16} & 90.33\wsd{0.26}       & \textbf{88.81}\wsd{0.14}  \\ \bottomrule
\end{tabular}
\caption{Negative log likelihood in the MNIST experiment. \textbf{Boldface} indicates the best result.}\label{tbl:riem-vae-mnist}
    \vspace{-0.7em}
\end{table}

The results are summarized in Table~\ref{tbl:riem-vae-mnist}. We can see that the implicit hyperspherical VAE trained with our method outperforms all other baselines. Interestingly, the explicit hyperspherical VAE could not match the performance of Euclidean VAE in higher dimensions. This is also observed in \citet{davidson2018hyperspherical}, who (incorrectly) conjectured that the hyperspherical prior is unsuitable in higher dimensions. From our results, we can see that the problem actually lies in the flexibility of variational posteriors. 
Our method thus unleashes the potential of VAEs with manifold-valued priors, and might lead to improvements in downstream tasks.

\subsubsection{Hyperspherical WAEs}

We first evaluate our method on MNIST. We use the uniform distribution as $p(z)$, and choose cross entropy as the reconstruction error. 
We choose $n_z=8$. We use the encoder and decoder architecture in \citet{song2019sliced}; the architecture of the energy network is also similar to their work. We report the Frechet Inception Distance \citep[FID;][]{heusel2017gans}. 

As the choice of divergence measure in the WAE objective is arbitrary, there are several methods to train WAEs with manifold latent space: using the Jensen-Shannon divergence approximated with a GAN-like discriminator (WAE-GAN), and using the maximum mean discrepancy (MMD) divergence. We choose WAE-GAN as the baseline method, as it outperforms WAE-MMD in \citet{tolstikhin2017wasserstein}. To demonstrate the utility of hyperspherical priors, we also compare with models using normal priors. 

\begin{table}[htbp]
    \centering
    \vspace{-0.3em}
\begin{tabular}{ccc}
\toprule
Method & Euc.        & Sph.        \\ \midrule
WAE-GAN        & 24.59\wsd{1.16}       & 19.81\wsd{1.13}       \\
Ours & 23.80\wsd{1.06}       & \textbf{18.36}\wsd{0.73}       \\ \bottomrule
\end{tabular}
\caption{FID for WAEs on MNIST.}%
\label{tbl:mnist-wae}
    \vspace{-0.7em}
\end{table}

The FID scores are reported in Table~\ref{tbl:mnist-wae}. We can see that hyperspherical prior leads to better sample quality compared with Euclidean prior, and our method improves the training of WAEs. %

To demonstrate our method scales to higher dimensions, we also train hyperspherical WAEs on CIFAR-10 and CelebA, with larger $n_z$. We find that our method is comparable or better than WAE-GAN and WAE-MMD; see Appendix~\ref{suppl:highd}. 

\vspace{-0.3em}
\section{CONCLUSION}
\vspace{-0.3em}

We present a scalable approximation to a general family of learning objectives for unnormalized models, based on a new connection between these objectives and gradient flows. Our method can be applied to manifold density estimation and training implicit auto-encoders with manifold priors.

\vspace{-0.3em}
\section*{ACKNOWLEDGEMENT}
\vspace{-0.3em}

J.Z is the corresponding author. We thank Chang Liu and Jiaxin Shi for comments. This work was supported by the National Key Research and Development Program of China (No. 2017YFA0700904), NSFC Project (Nos. 61620106010, U1811461), Beijing NSF Project (No. L172037), Beijing Academy of Artificial Intelligence (BAAI), a grant from Tsinghua Institute for Guo Qiang, and the NVIDIA NVAIL Program with GPU/DGX Acceleration.

\bibliography{main}

\begin{thebibliography}{}

\bibitem[Barp et~al., 2019]{aless2019minimum}
Barp, A., Briol, F.-X., Duncan, A.~B., Girolami, M., and Mackey, L. (2019).
\newblock Minimum stein discrepancy estimators.

\bibitem[Boomsma et~al., 2008]{boomsma2008generative}
Boomsma, W., Mardia, K.~V., Taylor, C.~C., Ferkinghoff-Borg, J., Krogh, A., and
  Hamelryck, T. (2008).
\newblock A generative, probabilistic model of local protein structure.
\newblock {\em Proceedings of the National Academy of Sciences},
  105(26):8932--8937.

\bibitem[Byrne and Girolami, 2013]{byrne2013geodesic}
Byrne, S. and Girolami, M. (2013).
\newblock Geodesic monte carlo on embedded manifolds.
\newblock {\em Scandinavian Journal of Statistics}, 40(4):825--845.

\bibitem[Chen et~al., 2018]{chen2018unified}
Chen, C., Zhang, R., Wang, W., Li, B., and Chen, L. (2018).
\newblock A unified particle-optimization framework for scalable bayesian
  sampling.
\newblock {\em arXiv preprint arXiv:1805.11659}.

\bibitem[Davidson et~al., 2018]{davidson2018hyperspherical}
Davidson, T.~R., Falorsi, L., De~Cao, N., Kipf, T., and Tomczak, J.~M. (2018).
\newblock Hyperspherical variational auto-encoders.
\newblock {\em arXiv preprint arXiv:1804.00891}.

\bibitem[Davis and Sampson, 1986]{davis1986statistics}
Davis, J.~C. and Sampson, R.~J. (1986).
\newblock {\em Statistics and data analysis in geology}, volume 646.
\newblock Wiley New York et al.

\bibitem[Federer, 2014]{federer2014geometric}
Federer, H. (2014).
\newblock {\em Geometric measure theory}.
\newblock Springer.

\bibitem[Gorham et~al., 2019]{gorham19measuring}
Gorham, J., Duncan, A.~B., Vollmer, S., and Mackey, L. (2019).
\newblock Measuring sample quality with diffusions.
\newblock {\em Annals of Applied Probability}.

\bibitem[Gutmann and Hyv{\"a}rinen, 2010]{gutmann2010noise}
Gutmann, M. and Hyv{\"a}rinen, A. (2010).
\newblock Noise-contrastive estimation: A new estimation principle for
  unnormalized statistical models.
\newblock In {\em Proceedings of the Thirteenth International Conference on
  Artificial Intelligence and Statistics}, pages 297--304.

\bibitem[Heusel et~al., 2017]{heusel2017gans}
Heusel, M., Ramsauer, H., Unterthiner, T., Nessler, B., and Hochreiter, S.
  (2017).
\newblock Gans trained by a two time-scale update rule converge to a local nash
  equilibrium.
\newblock In {\em Advances in Neural Information Processing Systems}, pages
  6626--6637.

\bibitem[Hsu, 2002]{hsu2002stochastic}
Hsu, E.~P. (2002).
\newblock {\em Stochastic analysis on manifolds}, volume~38.
\newblock American Mathematical Soc.

\bibitem[Hsu, 2008]{hsu2008brief}
Hsu, E.~P. (2008).
\newblock A brief introduction to brownian motion on a riemannian manifold.
\newblock {\em lecture notes}.

\bibitem[Hutchinson, 1990]{hutchinson1990stochastic}
Hutchinson, M.~F. (1990).
\newblock A stochastic estimator of the trace of the influence matrix for
  laplacian smoothing splines.
\newblock {\em Communications in Statistics-Simulation and Computation},
  19(2):433--450.

\bibitem[Hyv{\"a}rinen, 2005]{hyvarinen2005estimation}
Hyv{\"a}rinen, A. (2005).
\newblock Estimation of non-normalized statistical models by score matching.
\newblock {\em Journal of Machine Learning Research}, 6(Apr):695--709.

\bibitem[Hyvarinen, 2007]{hyvarinen2007connections}
Hyvarinen, A. (2007).
\newblock Connections between score matching, contrastive divergence, and
  pseudolikelihood for continuous-valued variables.
\newblock {\em IEEE Transactions on neural networks}, 18(5):1529--1531.

\bibitem[Li and Turner, 2018]{li2018gradient}
Li, Y. and Turner, R.~E. (2018).
\newblock Gradient estimators for implicit models.
\newblock In {\em International Conference on Learning Representations}.

\bibitem[Liu et~al., 2016a]{liu2016stochastic}
Liu, C., Zhu, J., and Song, Y. (2016a).
\newblock Stochastic gradient geodesic mcmc methods.
\newblock In {\em Advances in Neural Information Processing Systems}, pages
  3009--3017.

\bibitem[Liu et~al., 2019a]{liu2019understandingA}
Liu, C., Zhuo, J., Cheng, P., Zhang, R., Zhu, J., and Carin, L. (2019a).
\newblock Understanding and accelerating particle-based variational inference.
\newblock In Chaudhuri, K. and Salakhutdinov, R., editors, {\em Proceedings of
  the 36th International Conference on Machine Learning}, volume~97 of {\em
  Proceedings of Machine Learning Research}, pages 4082--4092, Long Beach,
  California USA. PMLR.

\bibitem[Liu et~al., 2019b]{liu2019understanding}
Liu, C., Zhuo, J., and Zhu, J. (2019b).
\newblock Understanding mcmc dynamics as flows on the wasserstein space.
\newblock {\em arXiv preprint arXiv:1902.00282}.

\bibitem[Liu, 2017]{liu2017stein}
Liu, Q. (2017).
\newblock Stein variational gradient descent as gradient flow.
\newblock In {\em Advances in neural information processing systems}, pages
  3115--3123.

\bibitem[Liu et~al., 2016b]{liu2016kernelized}
Liu, Q., Lee, J., and Jordan, M. (2016b).
\newblock A kernelized stein discrepancy for goodness-of-fit tests.
\newblock In {\em International conference on machine learning}, pages
  276--284.

\bibitem[Liu and Wang, 2016]{liu2016stein}
Liu, Q. and Wang, D. (2016).
\newblock Stein variational gradient descent: A general purpose bayesian
  inference algorithm.
\newblock In {\em Advances in neural information processing systems}, pages
  2378--2386.

\bibitem[Liu and Wang, 2017]{liu2017learning}
Liu, Q. and Wang, D. (2017).
\newblock Learning deep energy models: Contrastive divergence vs. amortized
  mle.
\newblock {\em arXiv preprint arXiv:1707.00797}.

\bibitem[{Lu} et~al., 2019]{lu2019accelerating}
{Lu}, Y., {Lu}, J., and {Nolen}, J. (2019).
\newblock {Accelerating Langevin Sampling with Birth-death}.
\newblock {\em arXiv e-prints}, page arXiv:1905.09863.

\bibitem[Lyu, 2009]{lyu2009interpretation}
Lyu, S. (2009).
\newblock Interpretation and generalization of score matching.
\newblock In {\em Proceedings of the Twenty-Fifth Conference on Uncertainty in
  Artificial Intelligence}, pages 359--366. AUAI Press.

\bibitem[Ma et~al., 2015]{ma2015complete}
Ma, Y.-A., Chen, T., and Fox, E. (2015).
\newblock A complete recipe for stochastic gradient mcmc.
\newblock In {\em Advances in Neural Information Processing Systems}, pages
  2917--2925.

\bibitem[Mardia et~al., 2016]{mardia2016score}
Mardia, K.~V., Kent, J.~T., and Laha, A.~K. (2016).
\newblock Score matching estimators for directional distributions.
\newblock {\em arXiv preprint arXiv:1604.08470}.

\bibitem[Mathieu et~al., 2019]{mathieu2019hierarchical}
Mathieu, E., Lan, C.~L., Maddison, C.~J., Tomioka, R., and Teh, Y.~W. (2019).
\newblock Continuous hierarchical representations with poincar\'e variational
  auto-encoders.
\newblock In {\em Advances in neural information processing systems}.

\bibitem[Miyato et~al., 2018]{miyato2018spectral}
Miyato, T., Kataoka, T., Koyama, M., and Yoshida, Y. (2018).
\newblock Spectral normalization for generative adversarial networks.
\newblock In {\em International Conference on Learning Representations}.

\bibitem[Movellan, 2007]{movellan2007minimum}
Movellan, J.~R. (2007).
\newblock A minimun velocity approach to learning.
\newblock {\em unpublished}.

\bibitem[Neal, 2001]{neal2001annealed}
Neal, R.~M. (2001).
\newblock Annealed importance sampling.
\newblock {\em Statistics and computing}, 11(2):125--139.

\bibitem[Otto, 2001]{otto2001geometry}
Otto, F. (2001).
\newblock The geometry of dissipative evolution equations: the porous medium
  equation.

\bibitem[Ovinnikov, 2019]{anonymous2020poincar}
Ovinnikov, I. (2019).
\newblock Poincar{\'{e}} wasserstein autoencoder.
\newblock {\em arXiv preprint arXiv:1901.01427}.

\bibitem[Ramachandran et~al., 2017]{ramachandran2017searching}
Ramachandran, P., Zoph, B., and Le, Q.~V. (2017).
\newblock Searching for activation functions.
\newblock {\em arXiv preprint arXiv:1710.05941}.

\bibitem[Raphan and Simoncelli, 2011]{raphan2011least}
Raphan, M. and Simoncelli, E.~P. (2011).
\newblock Least squares estimation without priors or supervision.
\newblock {\em Neural computation}, 23(2):374--420.

\bibitem[Ruiz and Titsias, 2019]{ruiz2019contrastive}
Ruiz, F.~J. and Titsias, M.~K. (2019).
\newblock A contrastive divergence for combining variational inference and
  mcmc.
\newblock {\em arXiv preprint arXiv:1905.04062}.

\bibitem[Saremi et~al., 2018]{saremi2018deep}
Saremi, S., Mehrjou, A., Sch{\"o}lkopf, B., and Hyv{\"a}rinen, A. (2018).
\newblock Deep energy estimator networks.
\newblock {\em arXiv preprint arXiv:1805.08306}.

\bibitem[Shi et~al., 2018]{shi2018spectral}
Shi, J., Sun, S., and Zhu, J. (2018).
\newblock A spectral approach to gradient estimation for implicit
  distributions.
\newblock In {\em Proceedings of the 35th International Conference on Machine
  Learning}, pages 4651--4660.

\bibitem[Sohl-Dickstein et~al., 2011]{sohl2011minimum}
Sohl-Dickstein, J., Battaglino, P., and DeWeese, M.~R. (2011).
\newblock Minimum probability flow learning.
\newblock In {\em Proceedings of the 28th International Conference on
  International Conference on Machine Learning}, pages 905--912. Omnipress.

\bibitem[Song et~al., 2019]{song2019sliced}
Song, Y., Garg, S., Shi, J., and Ermon, S. (2019).
\newblock Sliced score matching: A scalable approach to density and score
  estimation.
\newblock {\em arXiv preprint arXiv:1905.07088}.

\bibitem[Sriperumbudur et~al., 2017]{sriperumbudur2017density}
Sriperumbudur, B., Fukumizu, K., Gretton, A., Hyv{\"a}rinen, A., and Kumar, R.
  (2017).
\newblock Density estimation in infinite dimensional exponential families.
\newblock {\em The Journal of Machine Learning Research}, 18(1):1830--1888.

\bibitem[Srivastava et~al., 2007]{srivastava2007riemannian}
Srivastava, A., Jermyn, I., and Joshi, S. (2007).
\newblock Riemannian analysis of probability density functions with
  applications in vision.
\newblock In {\em 2007 IEEE Conference on Computer Vision and Pattern
  Recognition}, pages 1--8. IEEE.

\bibitem[Taghvaei and Mehta, 2019]{taghvaei19accelerated}
Taghvaei, A. and Mehta, P. (2019).
\newblock Accelerated flow for probability distributions.
\newblock In Chaudhuri, K. and Salakhutdinov, R., editors, {\em Proceedings of
  the 36th International Conference on Machine Learning}, volume~97 of {\em
  Proceedings of Machine Learning Research}, pages 6076--6085, Long Beach,
  California, USA. PMLR.

\bibitem[Tolstikhin et~al., 2017]{tolstikhin2017wasserstein}
Tolstikhin, I., Bousquet, O., Gelly, S., and Schoelkopf, B. (2017).
\newblock Wasserstein auto-encoders.
\newblock {\em arXiv preprint arXiv:1711.01558}.

\bibitem[Villani, 2008]{villani2008optimal}
Villani, C. (2008).
\newblock {\em Optimal Transport: Old and New}.
\newblock Grundlehren der mathematischen Wissenschaften. Springer Berlin
  Heidelberg.

\bibitem[Vincent, 2011]{vincent2011connection}
Vincent, P. (2011).
\newblock A connection between score matching and denoising autoencoders.
\newblock {\em Neural computation}, 23(7):1661--1674.

\bibitem[Wenliang et~al., 2018]{wenliang2018learning}
Wenliang, L., Sutherland, D., Strathmann, H., and Gretton, A. (2018).
\newblock Learning deep kernels for exponential family densities.
\newblock {\em arXiv preprint arXiv:1811.08357}.

\bibitem[Wenliang et~al., 2019]{li2019learning}
Wenliang, L., Sutherland, D., Strathmann, H., and Gretton, A. (2019).
\newblock Learning deep kernels for exponential family densities.
\newblock In Chaudhuri, K. and Salakhutdinov, R., editors, {\em Proceedings of
  the 36th International Conference on Machine Learning}, volume~97 of {\em
  Proceedings of Machine Learning Research}, pages 6737--6746, Long Beach,
  California, USA. PMLR.

\bibitem[Wu et~al., 2016]{wu2016quantitative}
Wu, Y., Burda, Y., Salakhutdinov, R., and Grosse, R. (2016).
\newblock On the quantitative analysis of decoder-based generative models.
\newblock {\em arXiv preprint arXiv:1611.04273}.

\bibitem[Xifara et~al., 2014]{xifara2014langevin}
Xifara, T., Sherlock, C., Livingstone, S., Byrne, S., and Girolami, M. (2014).
\newblock Langevin diffusions and the metropolis-adjusted langevin algorithm.
\newblock {\em Statistics \& Probability Letters}, 91:14--19.

\bibitem[Zhang et~al., 2018]{zhang2018stochastic}
Zhang, J., Zhang, R., and Chen, C. (2018).
\newblock Stochastic particle-optimization sampling and the non-asymptotic
  convergence theory.
\newblock {\em arXiv preprint arXiv:1809.01293}.

\end{thebibliography}
\def\harvardyearleft{(}
\def\harvardyearright{)}

\newpage

\onecolumn
\appendix

{\bf \Large Supplementary Material}

\section{Derivation of \eqref{eq:dsm-stoc-final-s}-\eqref{eq:dsm-stoc-final-e}}\label{suppl:deriv-dsm-variance}

Denote $\tilde{x}:=x+\sqrt{\epsilon}z$.

\begin{align}\label{eq:dsm-stoc}
\hat{L}_\mrm{dsm} 
&= \norm{x+\sqrt{\epsilon} z-\epsilon\nabla\cE(x+\sqrt{\epsilon} z)-x}^2 \\
&= \epsilon\norm{z}^2 + \epsilon^2\norm{\nabla\cE(\tilde{x})}^2 - 2\epsilon^{3/2}\bracket{z,\nabla\cE(\tilde{x})} \\
&= \epsilon\norm{z}^2 + \epsilon^2\norm{\nabla\cE(\tilde{x})}^2 - 2\epsilon^{3/2}\bracket{z,\nabla\cE(x)+(\nabla^2\cE(x))(\sqrt{\epsilon}z)+O(\epsilon)} \\
&= \underbrace{\epsilon^2\left(
    \norm{\nabla\cE(\tilde{x})}^2 - 2z^\top (\nabla^2\cE(x))z
\right)}_{A} + \underbrace{\epsilon\norm{z}^2-2\epsilon^{3/2}z^\top\nabla\cE(x)}_{B} + o(\epsilon^2),
\end{align}
Notice
\begin{equation*}
\EE_z(z^\top\nabla^2\cE(x)z) = \Delta\cE(x)
\end{equation*}
which is known as the Hutchinson's trick \citep{hutchinson1990stochastic}, so $\lim_{\epsilon\rightarrow 0}\epsilon^{-2}\EE(A)$ is two times the Fisher divergence $D_\mrm{F}(p|q)$. But $Var(B)=O(\epsilon^2)$, so as $\epsilon\rightarrow 0$, the rescaled estimator
$\epsilon^{-2}{\hat{L}_\mrm{dsm}}$
becomes unbiased with \emph{infinite variance}; and subtracting (B) from (A) results in a finite-variance estimator.

\section{On SPOS and MVL}

\newcommand{\Lsvec}{L_2(\rho\cX\rightarrow\RR^d)}
\newcommand{\Lssca}{L_2(\rho\cX\rightarrow\RR)}

\paragraph{Notations} In this section, let the parameter space be $d$-dimensional, and define $\Lsvec$ as the space of $d$-dimensional functions $\{f:\EE_{\rho(x)}\norm{f(x)}^2<\infty\}$. 

While in the main text, we identified the tangent space of $\Wass(\cX)$ as a subspace of $\Lsvec$ for clarity, here we use the equivalent definition $\mc{T}_\rho(\Wass(\cX)):=\{s\in\Lssca:\EE_{\rho}s=0\}$ following \citep{otto2001geometry}. The two definition are connected by the transform $s=-\nabla\cdot(\rho p)$ for $p\in\Lsvec$. Using the new definition, the differential of the KL divergence functional is then
$
(d\mrm{KL}_\phi)_\rho(s) := \int s(x) \log \frac{\rho(x)}{\phi(x)} dx.
$

\subsection{SPOS as Gradient Flow}\label{suppl:spos-gf}

In this section, we give a formal derivation of SPOS as the gradient flow of the KL divergence functional, with respect to a new metric.

Recall the SPOS sampler targeting distribution (with density) $\phi$ corresponds to the following density evolution:
\begin{align*}
    \partial_t\rho_t = -\nabla\cdot(\rho_t(x)\underbrace{(
        \phi_{\rho_t,\phi}^*(x) + 
        \alpha\nabla\log (\phi/\rho)
    )}_{\nu_t(x)})
\end{align*}
where $\alpha>0$ is a hyperparameter, and
$$
\phi_{\rho_t,\phi}^*(x)
:= \EE_{\rho_t(x')}(S_\phi\otimes k)(x',x)
:= \EE_{\rho_t(x')}[(\nabla_{x'}\log\phi(x'))k(x',x)+\nabla_{x'}k(x',x)]
$$
is the SVGD update direction \citep{liu2016stein,liu2017stein}.  
Fix $\rho$, define the integral operator
$$
K_\rho[f](x) := \EE_{\rho(x')} k(x',x)f(x),
$$
and define the tensor product operator $K_\rho^{\otimes d}:L^2(\cX\rightarrow \RR^d)\rightarrow L^2(\cX\rightarrow \RR^d)$ accordingly. Then the SVGD update direction satisfies
\begin{equation}\label{eq:svgd-dir-expr}
    \phi^*_{\rho,\phi} = K_\rho^{\otimes d}[\nabla\log(\phi/\rho)],
\end{equation}
which we will derive at the end of this subsection for completeness. Following \eqref{eq:svgd-dir-expr} we have
\begin{equation}\label{eq:spos-dir-expr}
    \nu_t(x) = (\alpha\mrm{Id}+K_\rho^{\otimes d})[\nabla\log(\phi/\rho)].
\end{equation}

The rest of our derivation follows \citep{otto2001geometry,liu2017stein}: 
consider the function space
$\cH_{\rho,\alpha}:=\{(\alpha\mrm{Id}+K_{\rho_t}^{\otimes d})[\nabla h]\},
$
where $h:\cX\rightarrow\RR$ is any square integrable and differentiable function. It connects to the tangent space of $\Wass(\cX)$ if we consider $s=-\nabla\cdot(\rho\tilde{p})$ for any $\tilde{p}\in\cH_{\rho,\alpha}$. 
Define on $\cH_{\rho,\alpha}$ the inner product
\begin{equation}\label{eq:tangent-space-metric}
\bracket{f,g}_{\cH_{\rho,\alpha}} := \bracket{f,(\alpha\mathrm{Id}+K_\rho^{\otimes d})^{-1}[g]}_{\Lsvec}.
\end{equation}
It then determines a Riemannian metric on the function space. 
For $\tilde{p}\in\cH_{\rho,\alpha}$ and $s=-\nabla\cdot(\rho\tilde{p})$, 
by \eqref{eq:spos-dir-expr} we have
$$
\bracket{\nu_t,\tilde{p}}_{\cH_{\rho,\alpha}} = 
\EE_{\rho_t(x)}\bracket{\nabla\log(\phi/\rho_t)(x),\tilde{p}(x)} = 
-\int \log\frac{\phi}{\rho_t}(\nabla\cdot(\tilde{p}\rho)) dx = -(d\mrm{KL}_\phi)(s),
$$
i.e. with respect to the metric \eqref{eq:tangent-space-metric}, SPOS is the gradient flow minimizing the KL divergence functional.

\paragraph{Derivation of \eqref{eq:svgd-dir-expr}}
let $(\lambda_i,\psi_i)_{i=1}^{\infty}$ be its eigendecomposition (i.e. the Mercer representation). For $j\in[d]$ let $\psi_{i,j}:=\psi_i\mbf{e}_j$ where $\{\mbf{e}_j\}_{j=1}^d$ is the coordinate basis in $\RR^d$, so $\{\lambda_i^{-1/2}\psi_{i,j}\}$ becomes an orthonormal basis in $\cH^{\otimes d}$. Now we calculate the coordinate of $\phi^*_{\rho,\phi}$ in this basis.
\begin{align}
    \bracket{\phi^*_{\rho,\phi}, \psi_{i,j}}_{L_2(\rho)}
&=  \EE_{\rho(x)}\EE_{\rho(x')}\bracket{(\nabla_{x'}\log\phi(x'))k(x',x)+\nabla_{x'}k(x',x), \psi_{i,j}(x)} \nonumber \\
&= \EE_{\rho(x')}\left[
       \bracket{\nabla_{x'}\log \phi(x'), (K_\rho[\psi_{i,j}])(x')} + 
       \nabla\cdot ((K_\rho[\psi_{i,j}])(x'))
    \right]  \nonumber\\
&=: \EE_{\rho(x')}[S_\phi(K_\rho[\psi_{i,j}])(x')].\label{eq:svgd-coord-near-done}
\end{align}
$S_\phi$ is known to satisfy the \emph{Stein's identity}
$$
\EE_{\rho} S_\rho(g) = 0
$$
for all $g\in \cH$. Thus, we can subtract $\EE_{\rho}S_\rho(K_\rho[\psi_{i,j}])$ from the right hand side of \eqref{eq:svgd-coord-near-done} without changing its value, and it becomes
\begin{align*}
     &\EE_{\rho(x')}[S_\phi(K_\rho[\psi_{i,j}])(x')] - \EE_{\rho(x')}[S_\rho(K_\rho[\psi_{i,j}])(x')]\\
=\:& \EE_{\rho(x')}\left[
       \left\bracket{\nabla_{x'}\log \frac{\phi(x')}{\rho(x')}, (K_\rho[\psi_{i,j}])(x')\right}
    \right]\\
=\:& \lambda_k \EE_{\rho(x')}\left[
       \left\bracket{\nabla_{x'}\log \frac{\phi(x')}{\rho(x')}, \psi_{i,j}(x')\right}
    \right].
\end{align*}
As the equality holds for all $i,k$, we completed the derivation of \eqref{eq:svgd-dir-expr}.

\subsection{MVL Objective Derived from SPOS}\label{suppl:mvl-spos-ksd}

By \eqref{eq:spos-dir-expr} and \eqref{eq:tangent-space-metric}, the MVL objective derived from SPOS is
$$
    \norm{\grad_\rho \mrm{KL}_\phi}^2_{\cH_{\rho,\alpha}} = 
    \bracket{\nabla\log(\phi/\rho_t),(\alpha\mrm{Id}+K^{\otimes d})\nabla\log(\phi/\rho_t)}_{\Lsvec}.
$$
In the right hand side above, the first term in the summation is the Fisher divergence, and the second is the kernelized Stein discrepancy \cite[Definition 3.2]{liu2016kernelized}. 

We note that a similar result for SVGD has been derived in \citep{liu2017learning}, and our derivations connect to the observation that Langevin dynamics can be viewed as SVGD with a Dirac function kernel (thus SPOS also corresponds to SVGD with generalized-function-valued kernels).

\section{Justification of the Use of Local Coordinates in \eqref{eq:loss-mvl-rld}}\label{suppl:local-coord-validity}

In this section, we prove in Proposition~\ref{prop:suppl-C-main} that the local coordinate representation lead to valid approximation to the MVL objective in the compact case. We also argue in Remark~\ref{rmk:suppl-C-numerics} that the use of local coordinate does not lead to numerical instability.

\begin{remark}
While a result more general than Proposition~\ref{prop:suppl-C-main} is likely attainable (e.g. by replacing compactness of $\cX$ with quadratic growth of the energy), this is out of the scope of our work; for our purpose, it is sufficient to note that the proposition covers manifolds like $S^n$, and the local coordinate issue will not exist in manifolds possessing a global chart, such as $H^n$. %
\end{remark}

\begin{lemma}\label{lem:C1} \textbf{(Theorem 3.6.1 in \citep{hsu2002stochastic})}
    For any manifold $\mc{M}$, $x\in \mc{M}$, and a normal neighborhood $B$ of $x$, %
    there exists constant $C>0$ such that the first exit time $\tau$ from $B$, of the Riemannian Brownian motion starting from $x$, satisfies
    $$
    P\left(\tau\le \frac{C}{L}\right)\le e^{-L/2}
    $$
    for any $L\ge 1$.
\end{lemma}

\begin{proposition}\label{prop:suppl-C-main}
    Assume the data manifold $\mc{X}$ is compact, and for all $\theta$, $\cE(\cdot;\theta)$ is in $C^1$. 
    Let $\tilde{L}_{\mrm{mvl\_rld}}$ be defined as in \eqref{eq:loss-mvl-rld}, $X_t$ following the true Riemannian Langevin dynamics targeting $q^{1/2}$. Then
    $$
    \frac{1}{2}\lim_{\epsilon\to 0}\EE(\tilde{L}_{\mrm{mvl\_rld}})=\left.\frac{d}{dt}\EE(\cE(X_t))\right|_{t=0},
    $$
    i.e. \eqref{eq:loss-mvl-rld} recovers true WMVL objective.
\end{proposition}
\newcommand{\PP}{\mathbb{P}}
\begin{proof}
By the tower property of conditional expectation, it suffices to prove the result when $P(X_0=x)=1$ for some $x$. 
Choose a normal neighborhood $B$ centered at $x$ such that $B$ is contained by our current chart, and has distance from the boundary of the chart bounded by some $\delta>0$. 
Let $C,\bar{\tau}$ be defined as in Lemma~\ref{lem:C1}. %
Recall the Riemannian LD is the sum of a drift and the Riemannian BM. Since $\mc{X}$ is compact and $\cE$ is in $C^1$, the drift term in the SDE will have norm bounded by some finite $C$. Thus the first exit time of the Riemannian LD is greater than $\min(\bar{\tau},\delta/C)=:\tau$. %

Let $X_t$ follow the true Riemannian LD, $\bar{X}_t=X_t$ when $t<\tau$, and be such that $\cE(\bar{X}_t)=0$ afterwards.\footnote{
This is conceptually similar to the standard augmentation used in stochastic process texts; from a algorithmic perspective it can be implemented by modifying the algorithm so that in the very unlikely event when $y^-$ escapes the chart, we return $0$ as the corresponding energy. 
We note that this is unnecessary for manifolds like $S^n$, since the charts can be extended to $\RR^d$ and hence $\tau=\infty$.
} 
By \citet{hsu2008brief}, until $\tau$, $\bar{X}_t$ follows the local coordinate representation of Riemannian LD \eqref{eq:riem-ld}, thus on the event $\{\epsilon\le \tau\}$, $\bar{X}_\epsilon$ would correspond to $y^-$ in \eqref{eq:riem-ld-again}. 
As $\mc{X}$ is compact, the continuous energy function $\cE$ is bounded by $|\cE(\cdot)|\leq A$ for some finite $A$. Then for sufficiently small $\epsilon$,
\begin{align*}
    \frac{1}{2}\EE(\tilde{L}_{\mrm{mvl\_rld}}) = \frac{\EE(\cE(\bar{X}_\epsilon)-\cE(X_0))}{\epsilon} &=
        \frac{\EE(\cE(X_\epsilon)-\cE(X_0))}{\epsilon}+\frac{\EE(\cE(\bar{X}_\epsilon)-\cE(X_\epsilon))}{\epsilon} \\
        &= \frac{\EE(\cE(X_\epsilon)-\cE(X_0))}{\epsilon}+\frac{\EE(-\cE(X_\epsilon)\mbf{1}_{\{\tau\le\epsilon\}})}{\epsilon}.
\end{align*}
In the above the first term converges to 
$
\left.\frac{d}{dt}\EE(\cE(X_t))\right|_{t=0}
$
as $\epsilon\rightarrow 0$, 
and 
$
\left|\frac{\EE(-\cE(X_\epsilon)\mbf{1}_{\{\tau\le\epsilon\}})}{\epsilon}\right|
\le \frac{A\PP(\tau\le\epsilon)}{\epsilon} = \frac{A\PP(\bar{\tau}\le\epsilon)}{\epsilon} \le \frac{Ae^{-C/2\epsilon}}{\epsilon}\to 0$ when $\epsilon\to 0$. Hence the proof is complete.
\end{proof}

\begin{remark}\label{rmk:suppl-C-numerics}
It is argued that simulating diffusion-based MCMC in local coordinates leads to numeric instabilities \citep{byrne2013geodesic,liu2016stochastic}. We stress that in our setting of approximating MVL objectives, this is not the case. The reason is that we only need to do a single step of MCMC, with arbitrarily small step-size. Therefore, we could use different step-size for each sample, based on the magnitude of $g$ and $\log q$ in their locations. We can also choose different local charts for each sample, which is justified by the proposition above.
\end{remark}

\section{Derivation of \eqref{eq:ent-est-to-score-est} in the Manifold Case}\label{suppl:deriv-ent-score-est}

In this section we derive \eqref{eq:ent-est-to-score-est}, when the latent-space distribution $q_\phi(z)$ is defined on a $p$-dimensional manifold embedded in some Euclidean space, and $H[q_\phi(z)]$ is the relative entropy w.r.t. the Hausdorff measure. The derivation is largely similar to the Euclidean case, and we only include it here for completeness. %

\eqref{eq:ent-est-to-score-est} holds because
\begin{align*}
\nabla_{\phi} \mathbb{H}\left[q_{\phi}( z)\right] 
&\stackrel{(i)}{=}
-\nabla_{ {\phi}} \mathbb{E}_{p(\epsilon)}\left[\log q_{ {\phi}}\left( {f}(\epsilon,{\phi})\right)\right] \\ 
&=
-\mathbb{E}_{ p( {\epsilon})}\left[\nabla_{ {\phi}} \log q_{ {\phi}}\left( {f}( {\epsilon},{\phi})\right)\right] \\
&=
-\mathbb{E}_{ p( {\epsilon})}\left[\left.\nabla_{ {\phi}} \log q_{ {\phi}}(z)\right|_{ {z}= {f}(\epsilon,\phi)}+\nabla_{{f}} \log q\left( {f}( {\epsilon}, \phi)\right) \nabla_{ {\phi}}  {f}(\epsilon,  \phi)\right] \\
&\stackrel{(ii)}{=}
-\mathbb{E}_{p( {\epsilon})}\left[\nabla_{ {z}} \log q_{\phi}(z) \nabla_{ {\phi}}  {f}(\epsilon,  \phi)\right],
\end{align*}
where (i) follows from Theorem 2.10.10 in \citet{federer2014geometric}, %
and (ii) follows from the same theorem as well as the fact that $\EE_{q_\phi(z)}[\nabla_\phi \log q_\phi(z)]=\nabla_\phi \int q_\phi(z)dz=0$. 

\section{Experiment Details and Additional Results}

{Code will be available at \url{https://github.com/thu-ml/wmvl}.}

\subsection{Synthetic Experiments}

\subsubsection{Experiment Details}\label{suppl:exp-details-smobj}

\paragraph{Experiment Details in Section~\ref{sec:exp-score-matching-obj}}
The (squared) bias is estimated as follows: denote the SSM estimator and ours as $\EE_{p(x)\mc{N}(\epsilon|0,1)}[L_F^{\mrm{ssm}}(x;\epsilon)]$ and $\EE_{p(x)\mc{N}(\epsilon|0,1)}[L_F^{\mrm{mvl}}(x;\epsilon)]$, respectively. One could verify that both methods estimate \eqref{eq:fisher-div-transformed}. 
Our estimate for the squared bias is now 
$
\frac{1}{K}\sum_{k=1}^K \left(\frac{1}{M}\sum_{j=1}^M (L_F^{\mrm{ssm}}(x^{(k)};\epsilon^{(j)})-L_F^{\mrm{mvl}}(x^{(k)};\epsilon^{(j)}))\right)^2.
$
where $x{(k)}\sim p(x),\epsilon^{(j)}\sim \mc{N}(0,1)$ are i.i.d. draws. The expectation of this estimate upper bounds the true squared bias by Cauchy's inequality, and the bias $\rightarrow 0$ as $K,M\rightarrow 0$. 
We choose $K=100,M=50000$ and plot the confidence interval. 
We also use these samples to estimate the variance of our estimator. 

For the model distribution $q$, we choose an EBM as stated in the main text. The energy of the model is parameterized as follows: we parameterize a $d$-dimensional vector $\psi(x;\theta)$ using a feed-forward network, then return $x^\top \psi(x;\theta)$ as the energy function. This is inspired by the ``score network'' parameterization in \citep{song2019sliced}; we note that this choice has little influence on the synthetic experiments (and is merely chosen here for consistency), but leads to improved performance in the AE experiments. 
Finally, $\psi(x;\theta)$ is parameterized with 2 hidden layers and Swish activation \citep{ramachandran2017searching}, and each layer has 100 units. We apply spectral normalization \citep{miyato2018spectral} to the intermediate layers. 
We train the EBM for 400 iterations with our approximation to the score matching objective, using a batch size of 200 and a learning rate of $4\times 10^{-3}$. The choice of training objective is arbitrary; changing it to sliced score matching does not lead to any notable difference, as is expected from this experiment.

The same procedure is applied to the denoising score matching estimator.

\paragraph{Experiment Details in Section~\ref{sec:exp-riem-sm}}
For this experiment, the data distribution is chosen as
$$
p(x) = 0.7 p_{vM}(x | (0,1),2) + 0.3 p_{vM}(x | (0.5,-0.5),3),
$$
where $p_{vM}$ is the von Mises density
$$
p_{vM}(x|\mu,\sigma) \propto e^{\frac{1}{\sigma^2}\cos(x-\mu)}.
$$ 
For the model distribution, the energy function is parameterized with a feed-forward network, using the same score-network-inspired parameterization as in the last experiment. The network uses tanh activation and has 2 hidden layers, each layer with 100 units. 

We generate 50,000 samples from $p(x)$ for training. We use full batch training and train for 6,000 iterations, using a learning rate of $5\times 10^{-4}$. The step-size hyperparameter in the MVL approximation is set to $10^{-5}$.

\subsubsection{On the Variance Problem in CD-1}\label{suppl:cd-cv}

To verify our control variate also solves the variance issue in CD-1, we train EBMs using CD-1 with varying step-size, with and without our control variate, and compare the score matching loss to EBMs trained with our method as well as sliced score matching. We use a separate experiment for CD-1 since it only estimates the gradient of the score matching loss.

\begin{figure}[htb]
    \centering
    \includegraphics[width=0.6\linewidth]{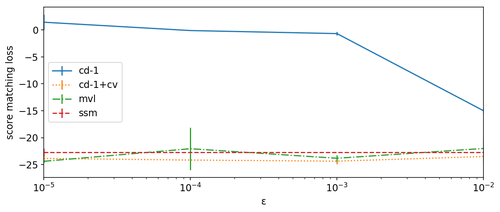}
    \caption{Score matching loss for different methods, with varying step-size. Lower is better.}
    \label{fig:smloss-cd-ebm}
\end{figure}

The score matching loss is calculated using SSM on training set, and averaged over 3 separate runs. We use the cosine dataset in \citep{wenliang2018learning}; the energy parameterization is the same as in Section~\ref{sec:exp-score-matching-obj}. The results are shown in Figure~\ref{fig:smloss-cd-ebm}. We can see that with the introduction of the control variate, CD-1 performs as well as other score matching methods.

\subsubsection{Learning EBMs on $S^2$}\label{suppl:s2}

\begin{figure}[htb]
    \centering
    \includegraphics[width=0.8\linewidth]{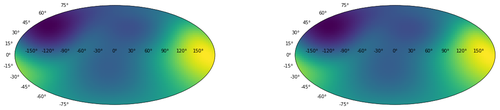}
    \caption{Mollweide projections of the ground truth (left) and learnt (right) energy functions on $S^2$.}
    \label{fig:ebm-s2}
\end{figure}

As a slightly more involved test case for our Riemannian score matching approximation, we consider learning EBMs on $S^2$. The target distribution is a mixture of 4 von-Mises-Fisher distributions. 
The ground truth and learnt energy functions are plotted in Figure~\ref{fig:ebm-s2}; we can see that our method leads to a good fit.

\subsection{Auto-Encoder Experiments}\label{suppl:exp-ae}

In all auto-encoder experiments, setup follows from \citep{song2019sliced} whenever possible. The only difference is that for score estimation, we parameterize the energy function, and use its gradient as the score estimate, as opposed to directly parameterizing the score function as done in \citep{song2019sliced}. This modification makes our method applicable; essentially, it corrects the score estimation in \citep{song2019sliced} so that it constitute a conservative field, which is a desirable property since score functions should be conservative.

For this reason, we re-implement all experiments for Euclidean-prior auto-encoders to ensure a fair comparison. The results are slightly worse than \citep{song2019sliced} for the VAE experiment, but significantly better for WAE experiments. 
It should be also noted that in the VAE experiment, our implicit hyperspherical VAE result is still better than the implicit Euclidean VAE result reported in \citep{song2019sliced}. 

\paragraph{VAE Experiment} The (conditional) energy function in this experiment is parameterized using the score-net-inspired method described in Appendix~\ref{suppl:exp-details-smobj}, with a feed-forward network. The network has 2 hidden layers, each with 256 hidden units. We use tanh activation for the network, and do not apply spectral normalization. 
When training the energy network, we add a L2 regularization term for the energy scale, with coefficient $10^{-4}$. The coefficient is determined by grid search on $\{10^{-3}, 10^{-4}, 10^{-5}\}$, using AIS-estimated likelihood on a heldout set created from the training set. 
The step-size of the MVL approximation is set to $10^{-3}$; we note that the performance is relatively insensitive w.r.t. the step-size inside the range of $[10^{-4},10^{-2}]$, as suggested by the synthetic experiment. Outside this range, using a smaller step-size makes the result worse, presumably due to floating point errors. 

For implicit models, the test likelihood is computed with annealed importance sampling, using 1,000 intermediate distributions, following \citep{song2019sliced}. The transition operator in AIS is HMC for Euclidean-space latents, and Riemannian LD for hyperspherical latents.

The training setup follows from \citep{song2019sliced}: for all methods, we train for 100,000 iterations using RMSProp use a batch size of 128, and a learning rate of $10^{-3}$. 

\paragraph{WAE Experiment on MNIST} For our method, the energy network is parameterized in the same way as in the VAE experiments. When training the energy network, we use a step-size of $10^{-3}$, and apply L2 regularization on the energy scale with coefficient $10^{-5}$. 
For the WAE-GAN baseline, we parameterize the GAN discriminator as a feed-forward network with 2 hidden layers, each with 256 units. %
We use tanh activation, and apply L2 regularization with coefficient $10^{-5}$. 
All models are trained for 200,000 iterations using RMSProp, using a batch size of 128, and a learning rate of $10^{-3}$. The Lagrange multiplier hyperparameter $\lambda$ in the WAE objective is fixed at $10$. 
FID scores are calculated using the implementation in \citep{heusel2017gans}.

\begin{figure}[hp]
  \centering
    \begin{subfigure}{.45\textwidth}
        \centering
        \includegraphics[width=\textwidth]{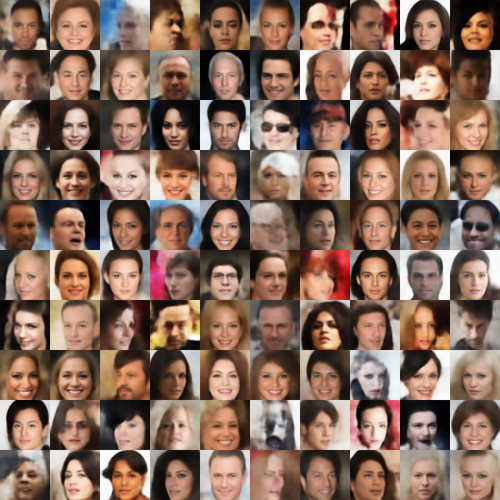}
        \caption{WAE-GAN, Hyperspherical Prior}
    \end{subfigure}
    \begin{subfigure}{.45\textwidth}
        \centering
        \includegraphics[width=\textwidth]{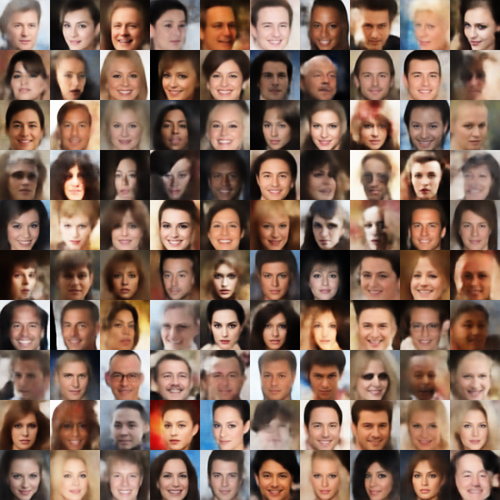}
        \caption{WAE-MVL, Hyperspherical Prior}
    \end{subfigure}
    \caption{Sampled generations in the WAE experiment on CelebA.}
    \label{fig:generation-wae-celeba}
\end{figure}

\paragraph{Sampled Generations in the Auto-encoder Experiments} See Figure~\ref{fig:generation-vae} - \ref{fig:generation-wae-celeba}.

\subsubsection{WAE Experiments in Higher Dimensions}\label{suppl:highd}

In this section, we present results of hyperspherical WAEs on CIFAR-10 and CelebA, with larger $n_z$. 

For CelebA we follow the setup in \citet{song2019sliced}: $n_z=32$, RMSProp, learning rate $10^{-4}$, train for 100,000 iterations. In addition, we apply spectral normalization and L2 regularization with coefficient $10^{-4}$. The step-size in the MVL approximation is set to $10^{-4}$. The FID scores, averaged over 5 runs, are $50.82\pm 0.50$ for our method and $51.20\pm 0.59$ for WAE-GAN.

\begin{figure}[htbp]
    \centering
    \includegraphics[width=0.4\linewidth]{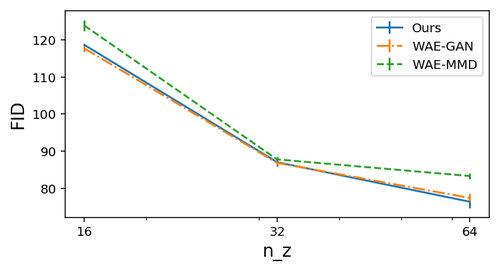}
    \includegraphics[width=0.4\linewidth]{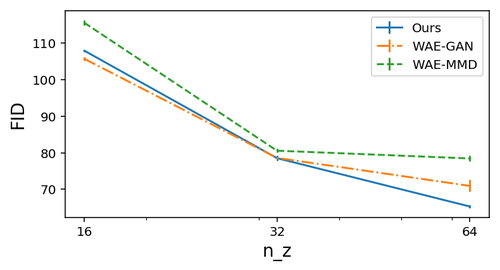}
    \caption{FID on CIFAR-10, with varying $n_z$. Left: after $10^5$ iterations; right: after $2\times 10^5$ iterations.}\label{fig:cifar}
\end{figure}

For CIFAR-10, we modify the auto-encoder architecture and remove one scaling block to account for its lower resolution. We do not use spectral normalization which leads to slightly worse results. The FID scores for varying $n_z$ are presented in Figure~\ref{fig:cifar}, where we can see our method compares favorably to all baselines.

\begin{figure}[hp]
    \centering
    \begin{subfigure}{.35\textwidth}
        \centering
        \includegraphics[width=\textwidth]{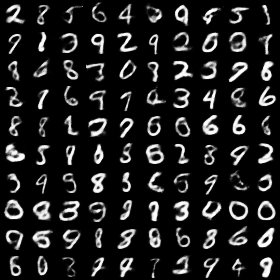}
        \caption{VAE, Euclidean Prior, $n_z=8$}
    \end{subfigure}
    \begin{subfigure}{.35\textwidth}
        \centering
        \includegraphics[width=\textwidth]{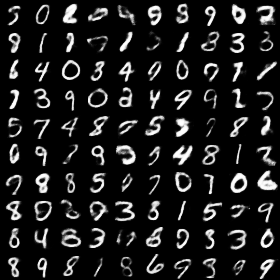}
        \caption{VAE, Hyperspherical Prior, $n_z=8$}
    \end{subfigure}
    \begin{subfigure}{.35\textwidth}
        \centering
        \includegraphics[width=\textwidth]{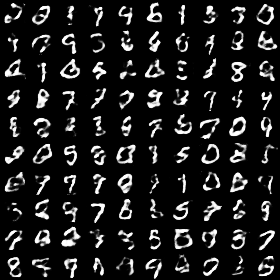}
        \caption{VAE, Euclidean Prior, $n_z=32$}
    \end{subfigure}
    \begin{subfigure}{.35\textwidth}
        \centering
        \includegraphics[width=\textwidth]{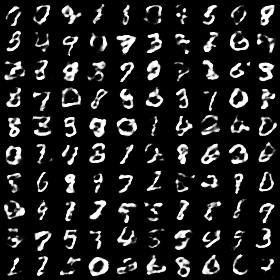}
        \caption{VAE, Hyperspherical Prior, $n_z=32$}
    \end{subfigure}
    \caption{Sampled generations of \textbf{implicit} VAEs.}
    \label{fig:generation-vae}
\end{figure}

\begin{figure}[hp]
    \centering
    \begin{subfigure}{.35\textwidth}
        \centering
        \includegraphics[width=\textwidth]{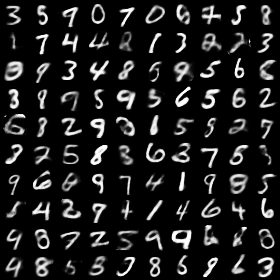}
        \caption{VAE, Euclidean Prior, $n_z=8$}
    \end{subfigure}
    \begin{subfigure}{.35\textwidth}
        \centering
        \includegraphics[width=\textwidth]{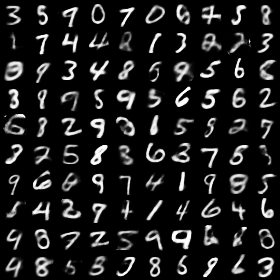}
        \caption{VAE, Hyperspherical Prior, $n_z=8$}
    \end{subfigure}
    \begin{subfigure}{.35\textwidth}
        \centering
        \includegraphics[width=\textwidth]{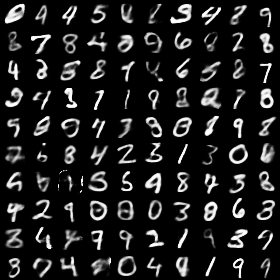}
        \caption{VAE, Euclidean Prior, $n_z=32$}
    \end{subfigure}
    \begin{subfigure}{.35\textwidth}
        \centering
        \includegraphics[width=\textwidth]{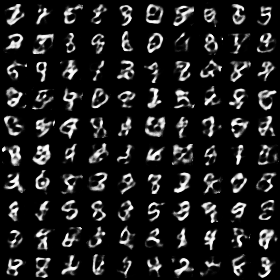}
        \caption{VAE, Hyperspherical Prior, $n_z=32$}
    \end{subfigure}
    \caption{Sampled generations of \textbf{explicit} VAEs.}
    \label{fig:generation-vae}
\end{figure}

\begin{figure}[hp]
    \centering
    \begin{subfigure}{.35\textwidth}
        \centering
        \includegraphics[width=\textwidth]{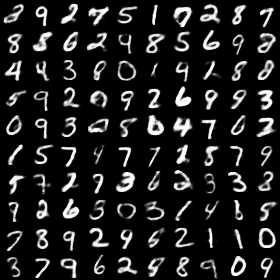}
        \caption{WAE-GAN, Euclidean Prior}
    \end{subfigure}
    \begin{subfigure}{.35\textwidth}
        \centering
        \includegraphics[width=\textwidth]{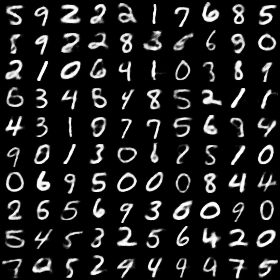}
        \caption{WAE-GAN, Hyperspherical Prior}
    \end{subfigure}
    \begin{subfigure}{.35\textwidth}
        \centering
        \includegraphics[width=\textwidth]{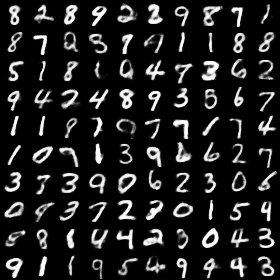}
        \caption{WAE-MVL, Euclidean Prior}
    \end{subfigure}
    \begin{subfigure}{.35\textwidth}
        \centering
        \includegraphics[width=\textwidth]{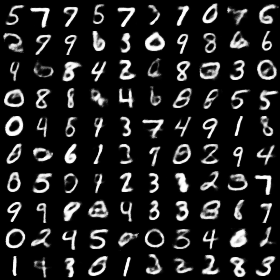}
        \caption{WAE-MVL, Hyperspherical Prior}
    \end{subfigure}
    \caption{Sampled generations in the WAE experiment on MNIST.}
    \label{fig:generation-wae}
\end{figure}

\end{document}